\documentclass[11pt,a4paper]{article}
\PassOptionsToPackage{hyphens}{url}\usepackage[hyperref]{emnlp-ijcnlp-2019}
\pdfoutput=1
\usepackage{times}
\usepackage{latexsym}
\usepackage{tikz-dependency}
\usepackage{graphicx}
\usepackage[hyphens]{url}
\usepackage{makecell}
\usepackage{multirow}
\usepackage{booktabs}
\usepackage[export]{adjustbox}
\usepackage{amsmath}
\usepackage{algorithm}
\usepackage[noend]{algpseudocode}
\usepackage{amsthm}   
\usepackage{amssymb} 
\usepackage{comment}

\aclfinalcopy 

\renewcommand{\vec}[1]{\mathbf{#1}}

\title{Towards Making a Dependency Parser See}

\author{Michalina Strzyz \qquad David Vilares \qquad Carlos G\'omez-Rodr\'iguez\\
  Universidade da Coru\~na, CITIC \\
  FASTPARSE Lab, LyS Research Group, Departamento de Computaci\'on \\
  Campus de Elvi\~na, s/n, 15071 A Coru\~na, Spain\\
  {\tt \{michalina.strzyz,david.vilares,carlos.gomez\}@udc.es}}

\date{}

\begin{document}
\maketitle
\begin{abstract}
   We explore whether it is possible to leverage eye-tracking data in an RNN dependency parser (for  English) when such information is only available during training -- i.e. no aggregated or token-level gaze features are used at inference time. To do so, we train a multitask learning model that parses sentences as sequence labeling and leverages gaze features as auxiliary tasks. Our method also learns to train from disjoint datasets, i.e. it
   can be used to test whether already collected gaze features 
   are useful to improve the performance on new \emph{non-gazed} annotated treebanks. Accuracy gains are modest but positive, showing the feasibility of the approach. It can serve as a first step towards architectures that can better leverage eye-tracking data or other complementary information available only for training sentences, possibly leading to improvements in syntactic parsing.

\end{abstract}

\section{Introduction}

Eye trackers and gaze features collected from them have been recently applied to natural language processing (\textsc{nlp}) tasks, such as part-of-speech tagging \cite{duffy1988lexical,nilsson2009learning,barrett2015reading}, named-entity recognition \cite{tokunaga2017eye} or readability \cite{gonzalez2018learning}. Eye-movement data has been also used for parsing. For example, \newcite{barrett2015using} rank discriminative features to predict syntactic categories (e.g. subject vs. object) and use them to improve a transition-based parser, trained on a structured perceptron with discrete features \cite{collins2002discriminative,zhang2011transition}. However, the experiments were carried out on a parallel toy treebank and the performance was relatively low. \newcite{lopopolo2019dependency} follow the inverse path, and use dependency parsing features to predict eye-regression during training, i.e. cases where the reader goes back to a word of the sentence.

In this context, how to retrieve and leverage eye-tracking data has become an active area of research in different \textsc{nlp} fields. 
Previous studies \cite{barrett2015using, barrett2016weakly} suggest that real-time eye-tracking data can be collected at inference time, so that \textit{token-level} gaze features are used during training but also at test time. However, even if in the near future every user has an eye tracker on top of their screen -- a scenario which is far from guaranteed, and raises privacy concerns \citep{Liebling2014} -- many running \textsc{nlp} applications that process data from various Internet sources will not expect to have any human being reading massive amounts of data. Other studies \citep{barrett2015reading,hollenstein2019entity} instead derive gaze features from the training set: in particular, they collect \textit{type-level} gaze features from the vocabulary in the training set; which are then aggregated to create a lookup table and used as a sort of precomputed gaze input when a given word in the test set matches an entry, otherwise, a token has an unknown gaze feature value. In this manner, the influence of gaze features on unseen data depends on the vocabulary encountered during training. For instance, \citet{hollenstein2019advancing} report that unknown tokens make up $41.09\%$ on new data for their research on named-entity recognition (\textsc{ner}) with eye-tracking data. To our knowledge, this kind of approaches have not been applied to syntactic parsing.

More in the line of our work, \citet{barrett2018sequence} is one of the few approaches that does not rely on the assumption of using gaze features as input. Instead, the human data is used as an inductive bias to guide the  attention weights in a recurrent neural network for sequence classification (used for tasks such as binary sentiment analysis a the sentence level). Moreover, it has been shown that both constituency \citep{VilaresMTL2019} and dependency \citep{StrVilGomACL2019} parsing can benefit from multi-task learning. Also related to our work with human data, \citet{gonzalez2017using} show that gaze features learned in a multi-task (\textsc{mtl}) setup can lead to improvements in readability assessment.

\paragraph{Contribution} In this work we leverage gaze learning for dependency parsing assuming that gaze features are likely to be only available during training -- and no gaze features are given, in any form, at inference time. Our approach learns eye-movement features as auxiliary tasks in a multitask framework where both parsing and gaze prediction are addressed as sequence labeling.
To test its effect on syntactic analysis, we experiment on parallel and disjoint datasets with dependency parsing and gaze annotations. The source code can be found at
\url{https://github.com/mstrise/dep2label-eye-tracking-data}.

\section{Methods and materials}

This section describes in detail the data and the models used in our work. 

\subsection{Data}

We will use both \emph{parallel} data (containing dependency and gaze annotations) and \emph{disjoint} data (where one dataset contains only parsing annotations and the other one just gaze data).

\paragraph{\emph{Parallel} data} The Dundee corpus~\cite{DundeeCorpus}  contains recordings of measurements for eye movements of ten English-speaking participants during reading $20$ newspaper articles from \textit{The Independent} making in total $2368$ unique sentences. This dataset was used as the starting point to create a corpus that fits the purpose of dependency parsing, the Dundee treebank \cite{DundeeTreebank}. This treebank augments sentences from the Dundee corpus with syntactic annotations according to the Universal Dependencies (UD) guidelines \cite{nivre2016universal}. We will use this treebank as the parallel data for our models. We split the data $(80-10-10)$ into training, development and test.
Sentences were randomly shuffled with assurance that the same sentence coming from the 10 participants is included only in one of the sets.\footnote{In addition we removed $40$ sentences ($4$ unique sentences read by the $10$ participants) that contained cases of incoherence in their syntactic dependency annotation resulting in $23640$ sentences in total ($2364$ unique sentences).}

\paragraph{\emph{Disjoint} data} Most of dependency treebanks do not contain gaze annotations. With this setup, in addition to the assumption that eye-movement data is available during training, we aim to show whether we improve the performance of a dependency parser with gaze-annotated data coming from a different corpus. We will use two datasets: one with dependency annotations and another one labeled with eye-movement data. For the former we use the Penn Treebank (PTB) \cite{ptb}, which contains sentences from \emph{The Wall Street Journal} annotated with phrase-structure trees. We convert it into Stanford Dependencies \cite{deMarneffe} and apply the standard splits for dependency parsing: sections $2-21$ for training, $22$ for development set and $23$ for testing, whereas PoS tags are predicted by the Stanford tagger \cite{tout}. For gaze-annotated data, we will employ the eye-movement annotations from the Dundee Treebank but using a different split $(90-10-0)$.\footnote{During testing we will be just testing the performance on the PTB test set.} 

\subsection{Gaze-averaged sequence labeling parsing}

During training we opt to emulate recent cognitive studies that suggest that human sentence understanding recruits the same brain regions for lexical and syntactic processing, consistent with a sequence-tagging-like process \cite{fedorenko2018word}, which is also a natural and intuitive way to learn to predict eye-movement information.
To do so, we rely on standard bidirectional long short-term memory networks (BILSTMs) \cite{hochreiter1997long,schuster1997bidirectional}. 

We denote LSTM$_\theta$($\vec{w}$) as the abstraction of a long short-term memory network, that processes an input sentence $w$=$[w_1,...,w_n]$ to generate an output of hidden representations $\vec{h}$=$[\vec{h}_1,...,\vec{h}_n]$. Thus, a BILSTM can be seen as BILSTM$_\theta(\vec{w})$ = LSTM$_\theta^l(\vec{w_{i:m}})\ \cdot$  LSTM$_\theta^r(\vec{w_{m:n}})$ = $\vec{h}^l \cdot \vec{h}^r$.\footnote{LSTM$^l$ processes the sentences from left-to-right and LSTM$^r$ from right-to-left. `$\cdot$' is the concatenation operator.} In this work we will stack two layers of BILSTMs before decoding the output.
BILSTMs are commonly used for sequence labeling tasks \cite{Reimers:2017:EMNLP}, where a word-level prediction can be generated for each token, by  adding a feed forward network to predict an output label at each time step, using a $\mathit{softmax}(\vec{W} \times \vec{h}_i + \vec{b})$. In this paper we propose to learn to leverage eye-movement data as a sequence labeling task. To jointly learn dependency parsing in a common framework, we also cast parsing as sequence labeling and learn both tasks in a multitask learning (MTL) setup \cite{caruana}. The main components are described in the following paragraphs.

\paragraph{Dependency parsing as sequence labeling} We proceed similarly to \citet{viable}. Given a linearization function  $F_{w}: T_{|w|} \rightarrow L^w$, for each word $w_i$, \citeauthor{viable}'s approach generates a label $l_i \in (o_i,p_i,d_i)$ that encodes the binary relationship between $w_i$ and its head, where: $d_i$ encodes the dependency relation, and the sub-pair ($o_i,p_i$) the index of such head term, with $o_i \in \mathbb{N}$ and $p_i \in P$ (a part-of-speech set). If $o_i$ is positive, then the head of $w_i$ is the $o_i$-th token to the right that has the part-of-speech tag $p_i$. If $o_i$ is negative, then the head of $w_i$ is the $|o_i|$-th token to the left whose PoS tag is $p_i$.  

This parser obtains similar results to competitive transition- and graph-based parsers such as BIST \cite{kiperwasser2016simple} and can be taken as a strong baseline to test the effect of eye-movement data for dependency parsing.

\paragraph{Gaze information}

In previous studies the choice of number of gaze features used in the experiments has varied, seemingly, depending on the \textsc{nlp} task of interest. For instance, \citet{barrett2016weakly} distinguish $31$ features (where $22$ are gaze features) for part-of-speech tagging while \citet{hollenstein2019entity} use $17$ features in the \textsc{ner} task. In another piece of work, $5$ gaze features are used for relation classification and sentiment analysis \citep{hollenstein2019advancing}. Finally, \citet{singh2016quantifying} use gaze features in order to automatically predict reading times for a new text. However, the model predicts only $4$ features that are then used as features for readability assessment.  

\begin{figure*}[ht]
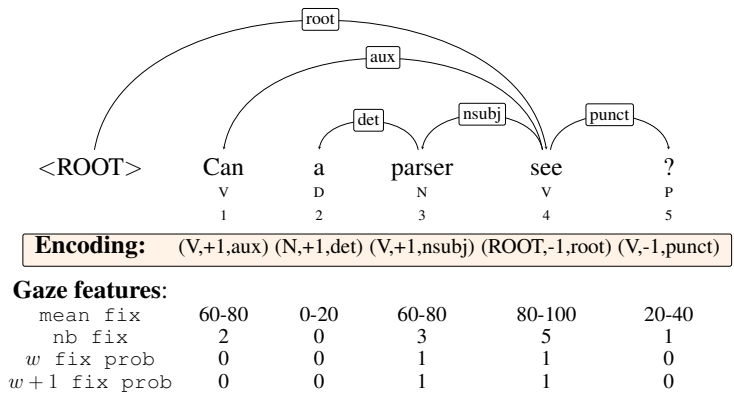

\centering
\scalebox{0.6}{
\begin{adjustbox}{width=\textwidth}
\begin{dependency}[label theme = default, edge vertical padding=3ex, arc edge, arc angle=80]
   \begin{deptext}[column sep=0.1em] 
     \Huge{$<$ROOT$>$} \&  \Huge{Can} \& \Huge{a} \& \Huge{parser} \& \Huge{see} \& \Huge{?}\\ \\
       \& \Large{V} \& \Large{D} \& \Large{N} \& \Large{V} \& \Large{P}\\  \\
       \& \Large{1} \& \Large{2} \& \Large{3} \& \Large{4} \& \Large{5}\\ \\ \\
      \Huge{\textbf{Encoding:}}   \& \huge{(V,+1,aux)} \& \huge{(N,+1,det)} \& \huge{(V,+1,nsubj)} \& \huge{(ROOT,-1,root)} \& \huge{(V,-1,punct)}\\ \\ \\ \\
      \Huge{\textbf{Gaze features}}:\\ \\
      \huge{\texttt{mean fix}} \& \huge{60-80} \& \huge{0-20} \& \huge{60-80} \& \huge{80-100} \& \huge{20-40} \\ \\
      \huge{\texttt{nb fix}} \& \huge{2} \&\huge{0} \& \huge{3} \& \huge{5} \&\huge{1}\\ \\
      \huge{\texttt{$w$ fix prob}} \& \huge{0} \&\huge{0} \& \huge{1} \& \huge{1} \&\huge{0}\\ \\
      \huge{\texttt{$w+1$ fix prob}} \& \huge{0} \&\huge{0} \& \huge{1} \& \huge{1} \&\huge{0}\\ 
   \end{deptext}
   
   \Huge{\depedge{5}{2}{aux}}
   \Huge{\depedge{4}{3}{det}}
   \Huge{\depedge{5}{4}{nsubj}}
   \Huge{ \depedge{1}{5}{root}}
    \Huge{\depedge{5}{6}{punct}}
     \wordgroup[group style={fill=orange!10, draw=black, inner sep=.8ex}]{8}{1}{6}{a0}
      
\end{dependency}
\end{adjustbox}
}
\caption{A dependency tree with its encoding and some of the corresponding gaze features.}
\label{fig:dependecyTree}
\end{figure*}

We have chosen $12$ gaze features and based on the previous work \citep{barrett2016weakly, hollenstein2019entity} we have divided them into $4$ groups. In particular, we explore the informativeness of the \textit{basic gaze features}: total fixation duration on a word $w$ (\texttt{total fix dur}), mean fixation duration on $w$ (\texttt{mean fix dur}, number fixations on $w$ (\texttt{$n$ fix}) and fixation probability on $w$ (\texttt{fix prob}). As \textit{early gaze features} we consider: first fixation duration on $w$ (\texttt{first fix dur}) and first pass duration (\texttt{first pass dur}), while as \textit{late gaze features}: number of re-fixations on $w$ ($n$ \texttt{re-fix}) and reread probability of $w$ (\texttt{reread prob}). We also take account of the neighboring words and treat them as \textit{context features}: fixation probability on the previous and next word ($w-1$ and $w+1$ \texttt{fix prob}) as well as fixation duration on the previous and next word ($w-1$ and $w+1$ \texttt{fix dur}). Figure \ref{fig:dependecyTree} depicts an exemplary (linearized) dependency tree with some of the gaze features used in the experiments. Following previous work \citep{hollenstein2019entity}, we discretize gaze features.\footnote{\texttt{Total fix dur}, \texttt{mean fix dur}, \texttt{first fix dur}, \texttt{first pass dur}, $w-1$ and $w+1$ \texttt{fix dur} are used as a discrete variable whose values are represented as percentile intervals with a bin size of 20, while for the other features we use their raw values. }

\paragraph{Multitask learning} We previously argued that gaze-annotated data is unlikely to be available at inference time. However, previous related work for dependency parsing usually assumes that such information will be collected during testing and fed as input features to the model \cite{barrett2015using,hollenstein2019entity}. What we propose instead is to leverage such information during training. To do so, we rely on MTL and auxiliary tasks. Our work focuses on exploiting the utility of gaze information using just a standard BILSTM, directly building on top of previous work of dependency parsing as sequence labeling \cite{viable}, and ignoring extra tools such as attention. In this line, a future possible solution could be to apply the approach by \citet{barrett2018sequence} to structured prediction and word-level classification. In their work they used human data as an inductive bias to update the attention weights of the network.

In our setup, dependency parsing as sequence labeling is addressed through two main tasks: one to predict the index of the head term, i.e. the sub-index $(o_i,p_i)$, and another one to predict the dependency relation ($d_i$). Eye-movement discrete labels are learned as auxiliary tasks. We use a hard-sharing architecture where the BILSTMs are fully shared and followed by an independent feed-forward layer (followed by a softmax) to predict the corresponding label for each of the tasks.
The main idea is that the eye-movement signal(s) will be back-propagated to update the weights of the shared BILSTM, building a model that latently encodes such information and helps dependency parsing.

For the parallel data setup, the model is trained in a standard way, and the cross-entropy loss is computed as $\mathcal{L} = \mathcal{L}_{(o,p)} + \mathcal{L}_{d} + \sum_{a} \beta_{a}*\mathcal{L}_{a}$. For the disjoint setup, during the training process for each batch we randomly pick all its samples from one of the treebanks (samples that have not been yet taken, either from the dependency parsing corpus or the gaze-annotated one) and run the network to make the predictions. Then, we back-propagate the loss for the outputs associated to that treebank and ignore the rest, i.e. $\mathcal{L} = \tau(\mathcal{L}_{(o,p)} + \mathcal{L}_{d}) + (1-\tau)\sum \beta_{aux}\mathcal{L}_{aux}$, where $\tau$ is a binary flag set to 1 when the batch contains dependency parsing samples, and to 0 otherwise.

For both setups we will test: (i) predicting \emph{one} of the gaze features as an auxiliary task (ii) predicting multiple gaze labels as multiple auxiliary tasks.

\section{Experiments}\label{section-experiments}

We conduct an explanatory study for the general viability of our method. We will compare results on the development and test set to verify whether the improvements are consistent. In particular, we carry out two experiments:

\begin{itemize}
\item Experiment 1 (on parallel data): Tests how eye-tracking data influences the performance when both gaze features and syntactic dependencies are extracted from the same data (in this case the Dundee treebank). The baseline is a model where no gaze features were used.

\item Experiment 2 (on disjoint data):

As previously said, we wish to test whether we can leverage existing gaze-annotated datasets to improve the performance in other non-gazed annotated dependency treebanks. Also, note that the Dundee treebank is relatively small ($\sim$2k sentences) and results from the previous Experiment can be not as representative as those obtained in larger corpora. In this setup, we train models using gaze information from the Dundee treebank and dependency relations from PTB, evaluating on the latter.\footnote{Note that the set of fine-tuned embeddings can be slightly different from that of Experiment 1, since words not occurring in Experiment 1 might be present in Experiment 2 due to the use of a different corpus.} The baseline is obtained by training a model only on the dependency relations retrieved from PTB.

\end{itemize}

\paragraph{Metrics} We evaluate the parsers with Unlabeled and Labeled Attachment Score (UAS and LAS) excluding punctuation (following the standard methodology for experiments on the PTB).

\subsection{Results}

\begin{table}[h]
\centering
\begin{adjustbox}{max width=\columnwidth}
\begin{tabular}{clllll}
\hline
\multicolumn{2}{c}{\multirow{2}{*}{Gaze features}} & \multicolumn{2}{c}{dev set} & \multicolumn{2}{c}{test set} \\
\multicolumn{2}{c}{} & \multicolumn{1}{c}{UAS} & \multicolumn{1}{c}{LAS} & UAS & LAS \\ \hline
\multicolumn{2}{l}{ \qquad \qquad \textit{baseline}} & 85.36 & 79.40  &84.37  &78.24 \\ \hline
\multirow{5}{*}{\textit{Basic}} & \texttt{total fix dur} & 85.34 & 79.35 & 84.06 & 77.44  \\
 & \texttt{mean fix dur} & 85.21 & 79.38 & 84.59 & \textbf{78.70}  \\
 & $n$ \texttt{fix} &  85.32 & 79.29 & 83.71 & 77.57   \\
 & \texttt{fix prob} &85.32 & 79.57 & 84.33 & 77.91 \\ \cline{2-6} 
 & \texttt{basic feats aux} & 85.36 & 79.57 & 83.86 & 77.75  \\\hline
\multirow{3}{*}{\textit{Early}} & \texttt{first fix dur} & 85.30 &79.46 &\textbf{84.64}  &78.57  \\
 & \texttt{first pass dur} & 85.50 & 79.49 & 84.55 & 78.39  \\ \cline{2-6} 
 & \texttt{early feats aux} & \textbf{85.61} & 79.57 & 84.37 & 78.11   \\\hline
\multirow{3}{*}{\textit{Late}} & $n$ \texttt{re-fix} & 85.52 & 79.25 & 83.86 & 77.91   \\
 & \texttt{reread prob} &85.34 & 79.57 & 83.86 & 77.37   \\ \cline{2-6} 
 & \texttt{late feats aux} &85.54 & 79.64 & 84.10 & 77.73  \\\hline
\multicolumn{1}{l}{\multirow{4}{*}{\textit{Context}}} & $w-1$ \texttt{fix prob} & 85.17 & 79.47 & 84.26 & 77.93   \\
\multicolumn{1}{l}{} & $w+1$ \texttt{fix prob} &85.36 & 79.07 & 84.24 & 78.06   \\
\multicolumn{1}{l}{} & $w-1$ \texttt{fix dur} & 85.43 & 79.68 & 84.50 & 77.95  \\
\multicolumn{1}{l}{} & $w+1$ \texttt{fix dur} &  85.39 & 79.53 & \textbf{84.64} & 78.30   \\ \cline{2-6} 
 \multicolumn{1}{l}{} &\texttt{context feats aux} & \textbf{85.61} & \textbf{79.72} &84.33&78.24  \\\hline
\end{tabular}
\end{adjustbox}
\caption{Impact of various gaze features as auxiliary task(s) on the score (UAS/LAS) of dependency parsing as the main task evaluated on Dundee treebank (parallel setup).}
\label{tab:allFeats}
\end{table}

\begin{table}[]
\centering
\begin{adjustbox}{max width=\columnwidth}
\begin{tabular}{clllll}
\hline
\multicolumn{2}{c}{\multirow{2}{*}{Gaze features}} & \multicolumn{2}{c}{dev set} & \multicolumn{2}{c}{test set} \\
\multicolumn{2}{c}{} & \multicolumn{1}{c}{UAS} & \multicolumn{1}{c}{LAS} & UAS & LAS \\ \hline
\multicolumn{2}{l}{ \qquad \qquad \textit{baseline}} & 93.98 & 91.67  &93.86 & 91.80  \\ \hline
\multirow{4}{*}{\textit{Basic}} & \texttt{total fix dur} & 93.94 & 91.60 & 93.99 & 91.92  \\
 & \texttt{mean fix dur} &  \textbf{94.12} & \textbf{91.84} & 93.95 & 91.82  \\
 & $n$ \texttt{fix} & 93.97 & 91.70 & 93.91 & 91.87   \\
 & \texttt{fix prob} & 93.98 & 91.71 & 93.99 & 91.93   \\ \cline{2-6} 
 & \texttt{basic feats aux} &94.00 & 91.69 & 93.84& 91.81 \\\hline
\multirow{2}{*}{\textit{Early}} & \texttt{first fix dur} &  94.07 & 91.81 & 93.87 & 91.80  \\
 & \texttt{first pass dur} & 93.93& 91.58 &93.79 &91.70   \\ \cline{2-6} 
 & \texttt{early feats aux} &94.04 & 91.78 & 93.96 & 91.88  \\\hline
\multirow{2}{*}{\textit{Late}} & $n$ \texttt{re-fix} & 94.01& 91.69 &93.87  &91.79   \\
 & \texttt{reread prob} & 94.03&91.74  &93.98  &91.89   \\\cline{2-6} 
 & \texttt{late feats aux} &93.98 & 91.58 & 93.92 & 91.90  \\\hline
\multicolumn{1}{l}{\multirow{4}{*}{\textit{Context}}} & $w-1$ \texttt{fix prob} & 94.02 & 91.65 & 93.95 & 91.93   \\
\multicolumn{1}{l}{} & $w+1$ \texttt{fix prob} & 93.88 & 91.61 & 93.89 & 91.82   \\
\multicolumn{1}{l}{} & $w-1$ \texttt{fix dur} &94.06 & 91.65 & 93.86 & 91.83  \\
\multicolumn{1}{l}{} & $w+1$ \texttt{fix dur} &93.91 & 91.69 & 93.89 & 91.84   \\ \cline{2-6} 
 & \texttt{context feats aux} &93.93 & 91.63 & \textbf{94.01} & \textbf{91.98}  \\\hline
\end{tabular}
\end{adjustbox}
\caption{Results for dependency parsing evaluated on PTB treebank with gaze features as auxiliary task(s) learned from the disjoint dataset: Dundee treebank.}
\label{tab:allFeatsPTB}
\end{table}

Table \ref{tab:allFeats} shows the results for Experiment 1 (on the Dundee treebank, i.e. on parallel data), suggesting that the degree of informativeness of the gaze features that a parser can leverage differ among each other.
In particular, we observe that the improvements across different models and the development and test sets are unstable.
The evaluation on the dev set shows that gaze features from \textit{early} and \textit{context} group used as multiple auxiliary tasks (\texttt{early} and \texttt{context feats aux}) modestly improve the model in comparison with the baseline. With respect to the results on the test set, the most informative features are: \texttt{mean fix dur}, which improves the
LAS by $+0.46$, and \texttt{first fix dur} by $+0.33$.

Table \ref{tab:allFeatsPTB} shows the results for Experiment 2 (using Dundee gaze data and PTB dependencies). Under this setup, the gains decrease in comparison with the results on the parallel setup. This could be partially caused by not using parallel data. The improvements seem to be more consistent between the development and test sets. This could be related to the fact that we use a larger (more representative) treebank.
When looking at the dev set scores, the most discriminative gaze feature is \texttt{mean fix dur} that increases LAS by $+0.17$ and \texttt{first fix dur} by $+0.14$.
On the other hand, evaluation on the test set shows that the most informative gaze features are from the \textit{context} group learned as multiple auxiliary tasks (\texttt{context feats aux}) and they improve the LAS score by $+0.18$, followed by \texttt{fix prob} and $w-1$ \texttt{fix prob} with $+0.13$, \texttt{total fix dur} with $+0.12$ and \texttt{late feats aux} with $+0.10$. 
Results from 
both datasets
suggest that grouping gaze features and treating them as multiple auxiliary tasks can improve the model's learning.

\paragraph{Discussion} The experiments show that our method can give moderate gains for dependency parsing when leveraging gaze information.\footnote{Nevertheless, our approach can be employed also to leverage other types of data than eye measurements, i.e. \textsc{ner} or chunking data.}
However, the experiments also show that there is room for improvement, especially coming from generalization capabilities across different treebanks. This also opens the question of whether a different architecture could better suit the purpose of leveraging the gaze information in a consistent way. In this context, a potential line of work could adapt human-attention approaches \cite{barrett2018sequence} for structured prediction and word-level classification, although it would come at a cost of speed for parsing as sequence labeling \cite{viable}.

\section{Conclusion}

This paper has explored how to leverage human data with a competitive dependency parser during training. This contrasts with most of previous work on parsing and gaze annotations, which worked under the assumption that eye-tracking data would be available at inference time.
We address this problem and propose a method of leveraging gaze features by using it as an auxiliary task both with and without parallel data. We obtain modest but positive improvements, which opens the question about how to increase the leverage of eye-tracking or other complementary data that is only available during training or comes from a different dataset. 

\section*{Acknowledgements}

This work has received funding from the
European Research Council (ERC), under the European Union's Horizon 2020 research and innovation programme (FASTPARSE, grant agreement No 714150), from the
ANSWER-ASAP project (TIN2017-85160-C2-1-R) from MINECO, and from Xunta de Galicia (ED431B 2017/01).

\bibliography{emnlp-ijcnlp-2019}
\bibliographystyle{acl_natbib}
\clearpage
\appendix
\section{Model parameters}\label{appendix-training-configuration}

Table \ref{tab:hyper} shows the hyperparameters used to train the models. We have trained each model up to 100 iterations and kept the one with the highest \textsc{las} score on the development set. The models were optimized with Stochastic Gradient Descent (SGD) with a batch size of 8. During multitask learning, dependency parsing as the main task was weighted $1.0$ while the gaze data treated as auxiliary task had a weight of $0.1$.

\begin{table}[hbtp]
\centering
\begin{adjustbox}{max width=\columnwidth}
\begin{tabular}{@{}ll@{}}
\toprule
Initial learning rate & 0.02 \\
Time-based learning rate decay & 0.05 \\
Momentum & 0.9 \\
Dropout & 0.5 \\ \midrule
\textsc{Dimensions} &  \\
Word embedding & 100 \\
Char embedding & 30 \\
Self-defined features & 20 \\
Word hidden vector & 800 \\
Character hidden vector & 50 \\ \bottomrule
\end{tabular}
\end{adjustbox}
\caption{Model hyperparameters. }
\label{tab:hyper}
\end{table}

\end{document}